\documentclass[letterpaper, 10 pt, conference]{ieeeconf}
\IEEEoverridecommandlockouts
\usepackage[bookmarks=true,hidelinks]{hyperref}
\usepackage[utf8]{inputenc}
\usepackage{amsmath,amsfonts}
\usepackage{algorithmic}
\usepackage{amssymb}%
\usepackage[ruled,vlined]{algorithm2e}
\usepackage{array}
\usepackage{graphicx}
\usepackage{csvsimple}
\usepackage{textcomp}
\usepackage{newtxtext}
\usepackage{stfloats}
\usepackage{url}
\usepackage{booktabs}
\usepackage{verbatim}
\graphicspath{{Figures/}}
\usepackage{orcidlink} 
\usepackage{cleveref}
\usepackage{multirow}

\def\BibTeX{{\rm B\kern-.05em{\sc i\kern-.025em b}\kern-.08em
    T\kern-.1667em\lower.7ex\hbox{E}\kern-.125emX}}
\usepackage{balance}

\begin{document}
%

\title{\LARGE \bf Robust Autonomous Control of a Magnetic Millirobot in In Vitro Cardiac Flow

\thanks{This work was funded by the National Science Foundation’s Foundational Research in Robotics CAREER program under award number 2144348.}%
\thanks{$^1$Laboratory for Computational Sensing and Robotics, Johns Hopkins University, Baltimore, MD 21218, USA. Emails:{\tt\small \{abhatt27, xchen254, axel\}@jhu.edu}}%
\thanks{$^2$Division of Magnetic Manipulation and Particle Research, Weinberg Medical Physics, Inc., Rockville, MD 20852, USA. Email:{\tt\small lamar.mair@gmail.com}}%
\thanks{$^3$Department of Mechanical Engineering, University of Maryland, College Park, MD 20742, USA. Emails:{\tt\small \{sraval, yancy\}@umd.edu}}%
\thanks{\textsuperscript{\textasteriskcentered}Corresponding Authors}%
}%

\author{Anuruddha Bhattacharjee$^{1}$\textsuperscript{\textasteriskcentered}\orcidlink{0000-0002-4053-4029}, Xinhao Chen$^{1}$\orcidlink{0009-0007-7381-4127}, Lamar O. Mair$^{2}$\orcidlink{0000-0001-9459-3932}, Suraj Raval$^{3}$\orcidlink{0000-0003-2889-6841}, \\Yancy Diaz-Mercado$^{3}$\orcidlink{0000-0003-0288-0112}, Axel Krieger$^{1}$\textsuperscript{\textasteriskcentered}\orcidlink{0000-0001-8169-075X}}

\maketitle

\begin{abstract}
Untethered magnetic millirobots offer significant potential for minimally invasive cardiac therapies; however, achieving reliable autonomous control in pulsatile cardiac flow remains challenging. This work presents a vision-guided control framework enabling precise autonomous navigation of a magnetic millirobot in an \textit{in vitro} heart phantom under physiologically relevant flow conditions. The system integrates UNet-based localization, A* path planning, and a sliding mode controller with a disturbance observer (SMC-DOB) designed for multi-coil electromagnetic actuation. Although drag forces are estimated using steady-state CFD simulations, the controller compensates for transient pulsatile disturbances during closed-loop operation. In static fluid, the SMC-DOB achieved sub-millimeter accuracy (root-mean-square error, RMSE = 0.49 mm), outperforming PID and MPC baselines. Under moderate pulsatile flow (7 cm/s peak, 20 cP), it reduced RMSE by 37\% and peak error by 2.4$\times$ compared to PID. It further maintained RMSE below 2 mm (0.27 body lengths) under elevated pulsatile flow (10 cm/s peak, 20 cP) and under low-viscosity conditions (4.3 cP, 7 cm/s peak), where baseline controllers exhibited unstable or failed tracking. These results demonstrate robust closed-loop magnetic control under time-varying cardiac flow disturbances and support the feasibility of autonomous millirobot navigation for targeted drug delivery.
\end{abstract}

\section{Introduction}
\begin{figure}[t]
\centering
\includegraphics[width=0.65\linewidth]{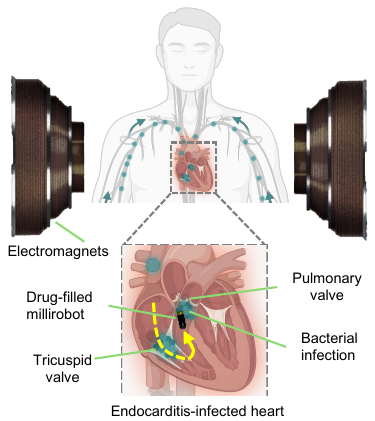}
\caption{Conceptual illustration of targeted cardiac drug delivery using magnetically actuated millirobot to treat infective endocarditis.}
\label{fig:application}
\vspace{-0.5\baselineskip}
\end{figure}

\noindent
Magnetically actuated milli-/microrobots are emerging as promising tools for targeted therapies in the cardiovascular system~\cite{iacovacci2024medical}. External magnetic fields enable wireless steering through narrow and tortuous anatomical pathways, such as blood vessels and cardiac chambers, with high precision~\cite{sitti2015biomedical}. Prior studies have demonstrated upstream motion against blood flow for drug delivery using microrollers~\cite{alapan2020multifunctional}, swimming in whole blood~\cite{wu2025femtosecond}, and autonomous exploration of vascular networks using magnetic microrobot swarms~\cite{du2025active}. These advances highlight the potential of magnetic robots for treating infective endocarditis~\cite{baddour2015infective}, in which bacteria infect cardiac valve regions (Fig.~\ref{fig:application}), necessitating precise, localized therapeutic delivery deep within the heart to achieve high drug concentrations at the infection site.

However, achieving reliable autonomous navigation in cardiac environments remains challenging. Blood flow inside the heart is fast, pulsatile, and spatially varying~\cite{doutel2021hemodynamics}, creating strong disturbances and model uncertainties that can destabilize magnetic actuation. Several closed-loop control strategies have been explored to improve robustness under flow. Sliding mode control (SMC) with a disturbance observer (DOB) has been shown to improve 3D path following for helical microrobots under strong flow disturbances and actuator saturation~\cite{qi2024robust}, while extended-state-observer SMC improved tracking in vascular phantoms~\cite{lu2022robust}. A double-loop SMC framework with an inner DOB has also been developed for electromagnetic actuator systems, enabling robust position and speed tracking under nonlinear model uncertainties~\cite{lee2025double}. In parallel, learning-based methods, including reinforcement learning~\cite{abbasi2024autonomous}, hierarchical deep reinforcement learning~\cite{yang2022hierarchical}, learning-enhanced model predictive control~\cite{zhou2025model}, adaptive learning–based control~\cite{liu2023adaptive}, and deep reinforcement learning–based navigation and flow rejection control~\cite{wang2024deep,cai2022deep}, have enabled adaptive path planning and control. These approaches have demonstrated improved tracking in static or tubular environments, including flow rejection under constant or sinusoidally varying flow with peak velocities of approximately 6--10~mm/s~\cite{cai2022deep} and precise tracking in static fluid without external flow disturbances~\cite{liu2023adaptive}. While these results represent important progress toward robust control, their experimental validation has largely been limited to relatively low flow velocities and simplified geometries.

Most existing controllers remain limited in their ability to operate reliably under realistic cardiac-like pulsatile flow conditions. Many model-based strategies assume steady or slowly varying drag, making them vulnerable to abrupt flow surges, recirculation, and transient disturbances present in cardiac chambers~\cite{miao2023magnetically}. Sliding mode controllers validated in static or vessel-like environments do not explicitly address rapidly varying pulsatile flow disturbances~\cite{lee2025double}, while data-driven approaches improve disturbance tolerance but often depend on training data collected under specific conditions and may not generalize across different flow regimes or complex cardiac geometries without retraining~\cite{jiang2022control}. Consequently, experimental demonstration of fully autonomous magnetic millirobot navigation under physiologically relevant pulsatile cardiac flow, involving time-varying disturbances and realistic cardiac geometry, remains limited.

This work develops a vision-guided autonomous control framework for a drug-filled magnetic millirobot in an \textit{in vitro} heart phantom under physiologically relevant pulsatile flow. The contributions are: 1) integration of UNet-based localization and A*-based trajectory planning for closed-loop navigation in cardiac phantom; 2) a sliding mode controller with disturbance observer (SMC-DOB) coupled with CFD-derived drag estimation to enable robust trajectory tracking under pulsatile disturbances; and 3) experimental validation of autonomous control under physiologically relevant pulsatile flow and viscosity variations, with comparative evaluation against baseline controllers.

\begin{figure*}[t]
\centering
\includegraphics[width=0.72\linewidth]{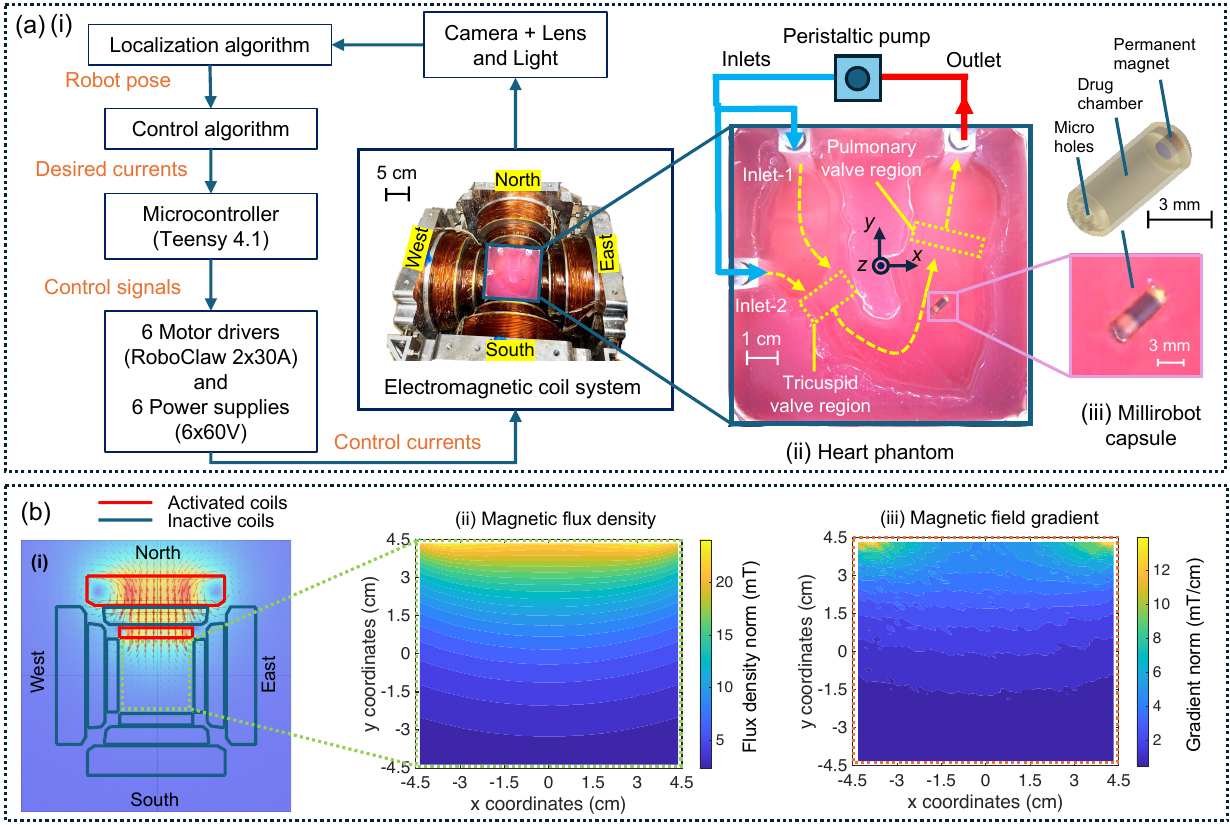}
\caption{(a) Experimental system: 
(i) Twelve-coil electromagnetic actuation setup surrounding the heart phantom, with a top-mounted camera for vision-based feedback control; 
(ii) Enlarged view of the heart phantom showing the internal flow circuit, where a peristaltic pump recreates right atrial and ventricular flow; 
(iii) Magnetic millirobot capsule containing an embedded permanent magnet, a drug chamber, and micro-holes for controlled drug release. 
(b) Finite element simulations of the electromagnetic system used to characterize magnetic flux density and field gradients under different coil activation patterns.}
\label{fig:magneticSystem}
\vspace{-1\baselineskip}
\end{figure*}

\section{System and Methods}\label{sec:system}

\subsection{Magnetic Robotic System}
\textbf{Electromagnetic actuation system:}
The electromagnetic coil system (Fig.~\ref{fig:magneticSystem}(a)(i)) enables wireless manipulation of magnetic millirobots within a central $10\times10\times10$~cm$^3$ workspace. The setup consists of twelve planar coils arranged as four units, each containing three concentric coils of different radii with axes aligned in the $x$–$y$ plane. Each coil is independently driven by six Roboclaw 2$\times$60~AHV motor drivers powered by 60~V supplies and controlled through a custom Python interface communicating with a Teensy microcontroller for pulse-width modulation. This configuration enables the generation of tunable magnetic field gradients for high-force actuation and precise positioning, with specifications comparable to previously reported multi-coil electromagnetic actuation systems~\cite{onder2023VORIACS}.

Finite element simulations (COMSOL Multiphysics) were performed to characterize the magnetic flux density and gradients under various coil activation patterns (Fig.~\ref{fig:magneticSystem}(b)). For example, activating the large and small north coils at 0.36~kW each produced a peak flux density of 24.05~mT and a gradient of 13.92~mT/cm (Fig.~\ref{fig:magneticSystem}(b)(ii)--(iii)). The simulated field maps were stored as look-up tables and used to compute millirobot forces and torques during closed-loop control. 

\textbf{Magnetic millirobot design:}
A capsule-type magnetic millirobot was developed for targeted drug delivery in the cardiac environment. Figure~\ref{fig:magneticSystem}(a)(iii) shows the robot filled with a dye solution (simulated drug) and its CAD model. The robot comprises a plastic body with an internal drug chamber, an embedded axially magnetized neodymium disk magnet (D0110-10, SuperMagnetMan), and a release cap for fluid discharge. The magnet has a 1~mm diameter and length (grade N50) and generates a numerically estimated pulling force of 0.37~mN under a 0.43~T/m magnetic field gradient based on FEA field simulations. The body components were fabricated using BIO resin on a microArch\textsuperscript{\textregistered} S140 resin 3D printer. The release cap contains seven circular holes (0.3~mm diameter) arranged in a ring with 0.6~mm center-to-center spacing, enabling passive drug release via diffusion into the surrounding fluid. The assembled capsule measures 2.80~mm in diameter and 7.40~mm in length, carrying up to 22~$\mu$L of liquid. These dimensions enable passage through cardiac structures such as the tricuspid valve and right ventricular outflow tract (27-33 mm~\cite{tsipis2022echocardiography}).

\textbf{Heart phantom with pulsatile flow:} A heart phantom was constructed as an \textit{in vitro} platform to evaluate millirobot navigation under physiologically relevant flow. The phantom was molded from 2\% (w/w) agarose gel to approximate cardiac tissue elasticity~\cite{tejo2022soft} and cast from a cross-sectional heart model using a silicone mold (Ecoflex\textsuperscript{TM} 00-20 Fast, 1:1 ratio). The resulting structure represents portions of the right atrium, right ventricle, and pulmonary trunk, including narrowed valve-like regions and curved channels near the tricuspid and pulmonary valves (Fig.~\ref{fig:application}, Fig.~\ref{fig:magneticSystem}(a)(ii)).

Pulsatile flow was generated using a peristaltic pump circulating glycerin–water mixtures through a closed-loop circuit connected to the phantom. Under the tested pump settings, the flow exhibited an approximate pulsation frequency of 3~Hz, within the physiological range observed under elevated cardiac activity. Flow rates of 180 and 250~mL/min produced peak velocities of approximately 7–10~cm/s, consistent with physiological right atrial flow during resting conditions~\cite{gabe1969measurement}. Fluid viscosity was adjusted by varying the glycerin–water ratio (40:60 to 65:35) to achieve 4.3 and 20~cP at 23$^\circ$C, representing near-healthy and elevated blood viscosity levels, respectively~\cite{taco2019association}. These conditions were used to evaluate controller robustness under varying fluid resistance and pulsatile disturbances.

\subsection{Modeling and State Estimation}

\textbf{Magnetic actuation dynamics:}\label{sec:magnetic_dynamics}
A magnetic field with spatial gradients exerts both force and torque on a robot containing an embedded permanent magnet. The magnetic force $\boldsymbol{F_{m}}$ and torque $\boldsymbol{T}$ are given by
\begin{equation} \label{eq:Basic_force_torque}
\boldsymbol{F_{m}} = \boldsymbol{m}^T \nabla\boldsymbol{B}, \quad
\boldsymbol{T} = \boldsymbol{m} \times \boldsymbol{B},
\end{equation}
where $\boldsymbol{m}$ is the magnetic dipole moment, $\nabla\boldsymbol{B}$ is the magnetic field gradient, and $\boldsymbol{B}$ is the magnetic flux density, which scales linearly with coil current.

The electromagnetic actuation system comprises twelve planar coils arranged for in-plane ($x$–$y$) control. In a low-viscosity fluid, the robot rapidly aligns with the field and requires only low field strength for translation; therefore, rotational dynamics are neglected, and eight coils are used for planar control. The mapping from coil currents to magnetic forces and field components is expressed as
\begin{equation}
\label{eq:Control_Equation}
\begin{bmatrix}
F_{mx} \\
F_{my} \\
\omega_o B_x \\
\omega_o B_y
\end{bmatrix}
=
\begin{bmatrix} 
\boldsymbol{m}^T
\begin{bmatrix}
\frac{\partial \widetilde{\boldsymbol{B_{1}}}}{\partial x}
\cdots
\frac{\partial \widetilde{\boldsymbol{B_{8}}}}{\partial x}
\end{bmatrix}
\\
\boldsymbol{m}^T
\begin{bmatrix}
\frac{\partial \widetilde{\boldsymbol{B_{1}}}}{\partial y}
\cdots
\frac{\partial \widetilde{\boldsymbol{B_{8}}}}{\partial y}
\end{bmatrix}
\\
\widetilde{B_{1}}_x \cdots \widetilde{B_{8}}_x
\\
\widetilde{B_{1}}_y \cdots \widetilde{B_{8}}_y
\end{bmatrix}
\begin{bmatrix}
I_1(t) \\
\vdots \\
I_{8}(t)
\end{bmatrix},
\end{equation}
where $\widetilde{\boldsymbol{B_{i}}}$ is the current-normalized magnetic field of coil $i$, $I_i$ is the input current, $B_x$ and $B_y$ are the magnetic field components, and $\omega_o$ is a unit conversion factor. This can be written compactly as $\boldsymbol{C} = \boldsymbol{A}\boldsymbol{I}$, where $\boldsymbol{A}$ is the current-to-field mapping matrix obtained from finite element analysis~\cite{chen2024mitigating}.

To compute the coil currents for a desired control vector $\boldsymbol{C}$, $\boldsymbol{A}$ is decomposed using singular value decomposition (SVD),
\begin{equation}
\label{eq:SVD}
\boldsymbol{A} =
\boldsymbol{U}\boldsymbol{\Sigma}\boldsymbol{W}^T,
\end{equation}
where $\boldsymbol{U}$ and $\boldsymbol{W}$ are orthogonal matrices, and $\boldsymbol{\Sigma}$ is diagonal with singular values. Then the input currents are obtained as
\begin{equation}
\label{eq:SVD_result}
\boldsymbol{I} =
\boldsymbol{W}
\boldsymbol{\Sigma}^+
\boldsymbol{U}^T
\boldsymbol{C},
\end{equation}
where $\boldsymbol{\Sigma}^+$ is the Moore–Penrose pseudoinverse.

During motion in a fluid, the robot experiences viscous drag,
\begin{equation}
\label{eq:Drag_force}
\begin{bmatrix}
\boldsymbol{F_{dx}} \\
\boldsymbol{F_{dy}}
\end{bmatrix}
=
\begin{bmatrix}
c_{t} \boldsymbol{V_{x}} \\
c_{t} \boldsymbol{V_{y}}
\end{bmatrix},
\end{equation}
where $\boldsymbol{V_{x}}$ and $\boldsymbol{V_{y}}$ are flow velocities. The translational damping coefficient for an ellipsoid is~\cite{salehizadeh2017two}
\begin{equation}
\label{eq:Drag_coefficient}
c_t =
\frac{2\pi \mu_d L}
{\log\!\left(\frac{L}{r}\right)},
\end{equation}
where $\mu_d$ being the dynamic viscosity of the fluid, and $L$ and $r$ are the robot length and radius.

\begin{figure}[t]
\centering
\includegraphics[width=0.95\linewidth]{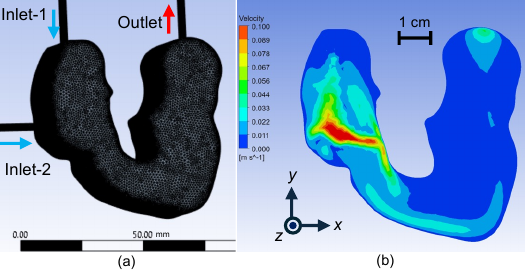}
\caption{(a) Computational mesh of the heart phantom model with two flow inlets and an outlet and (b) velocity magnitude contour plot for 4.3 cP fluid at the plane of robot control (i.e., $z$ = 8 mm, where $z$ = 0 mm is the plane passing through the centerline of the inlets and outlets) and $0.295$~cm/s flow speed at each inlet.}
\label{fig:cfd-plots}
\vspace{-1\baselineskip}
\end{figure}

\textbf{CFD-based flow field estimation:}
To estimate the flow field in the heart phantom, steady-state laminar flow was simulated in a computational fluid dynamics (CFD) software (ANSYS Fluent) using a simplified heart phantom geometry (Fig.~\ref{fig:cfd-plots}). The model consisted of two straight inlet channels ($D = 3$~mm, $L = 85$~mm) feeding a central chamber and a single outlet, with all lumen walls defined as no-slip boundaries. The computational mesh (Fig.~\ref{fig:cfd-plots}(a)) employed a patch-conforming tetrahedral core with prism-layer inflation along the walls ($\sim5.4\times10^5$ elements, 12 layers, first layer $0.015$~mm, growth rate 1.2). The working fluid was a Newtonian glycerin–water mixture ($\rho = 1113~\mathrm{kg\,m^{-3}}$, $\mu = 0.0043~\mathrm{Pa\,s}$), with velocity-inlet and pressure-outlet boundary conditions. A pressure-based SIMPLE solver with second-order schemes was used, and convergence was accepted when residuals fell below $10^{-6}$.

For inlet flow rates of $Q=250$ and $180$~mL/min, the average inlet velocities were $U\approx0.295$ and $0.212$~m/s, corresponding to Reynolds numbers $\mathrm{Re}\approx227$ and $164$. Using the laminar entrance length estimate $L_e \approx 0.05\,\mathrm{Re}\,D$~\cite{shah2014laminar}, the required entrance lengths (34 and 25~mm) were much shorter than the actual inlet length (85~mm), confirming fully developed inlet profiles. At higher viscosity ($\mu=0.020$~Pa$\cdot$s), $\mathrm{Re}$ dropped below 50 with similarly short entrance lengths. The converged steady-state velocity field on the actuation plane (Fig.~\ref{fig:cfd-plots}(b)) was exported, and only the in-plane ($x$-$y$) components $(V_x,V_y)$ were used for drag estimation and control. Figure~\ref{fig:cfd-plots}(b) shows the velocity contour at $250$~mL/min inlet flow rate, producing peak velocities up to 10~cm/s within the phantom near the inlet-2 region.

\textbf{Vision-based localization and pose estimation:}
Autonomous magnetic control requires real-time estimation of the millirobot’s position and orientation. We developed a vision-based localization framework using a modified UNet~\cite{ronneberger2015u} that performs both semantic segmentation and keypoint detection, enabling simultaneous estimation of the millirobot mask and its tip, centroid, and tail keypoints. Unlike prior work that used UNet solely for segmentation of magnetic needles in surgical settings~\cite{pryor2021localization}, this model was trained specifically for a drug-filled millirobot navigating inside the heart phantom.

A dataset of 2,005 labeled images was collected during millirobot experiments in the heart phantom. Segmentation masks were generated using a SAM2-based automated pipeline~\cite{ravi2024sam}, and keypoints were annotated manually. The model was trained using cross-entropy loss for segmentation and L1 loss for keypoints ($\mathcal{L} = \mathcal{L}_\text{CE} + 0.5\,\mathcal{L}_\text{L1}$) with Adam ($10^{-3}$), batch size 3, ReduceLROnPlateau scheduling, and early stopping. Augmentations included flips and $90^\circ$ rotations. During inference, cropped frames were resized to $572\times572$ and processed by the network. The centroid keypoint defined position, and the tip–tail vector defined orientation. This orientation information is required to align the applied magnetic field with the robot’s magnetic moment, ensuring that the generated magnetic force produces the intended translational motion. The model ran at 18~fps on an Nvidia GeForce RTX-4070-Ti and achieved \textbf{98.7\% precision}, \textbf{98.3\% recall}, and \textbf{97.1\% IoU}, enabling reliable real-time pose estimation under pulsatile flow.

\subsection{Planning and Control Framework}

\begin{figure}[t]
\centering
\includegraphics[width=0.6\linewidth]{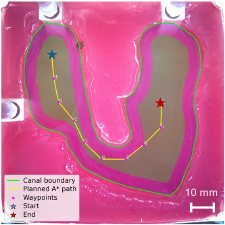}
\caption{A* planned trajectory (yellow) within the user-defined canal mask with 5 mm wall clearance and 50\% centerline bias. Start (blue star) and end (red star) points are connected through resampled waypoints (magenta), and the green spline indicates the user-selected canal boundary.}
\label{fig:astar-path}
\vspace{-1\baselineskip}
\end{figure}

\textbf{Path planning and trajectory generation:}
An image-based path planning pipeline using the A* algorithm was developed to generate reproducible navigation trajectories. A top-view image of the heart phantom (1020$\times$1020 pixels, $\approx$92$\times$92~mm$^2$) was used as the planning domain, where the navigable canal was manually outlined to create a binary mask. A Euclidean distance transform was applied to compute per-pixel wall clearance, and a 5~mm minimum clearance constraint defined the feasible core region. A cost map combining uniform motion cost with a clearance-based centerline bias ($w_\text{clear}=0.5$) was constructed within this region. User-selected start and end points were projected to the nearest feasible pixels, and an 8-connected A* search was performed with diagonal and straight step costs of $\sqrt{2}$ and 1, respectively. The resulting path was smoothed using a moving-average filter (window size $=7$) and resampled into 10 evenly spaced waypoints. These were converted to metric coordinates using the pixel-to-millimeter ratio and provided to the controller as the desired trajectory (Fig.~\ref{fig:astar-path}). This trajectory was computed offline and remained fixed during experiments, providing a consistent reference path for evaluating controller performance rather than dynamic replanning. The start and end points were selected to mimic navigation from the inlet toward the pulmonary valve region while traversing the tricuspid valve area, representing a clinically relevant path for targeted drug delivery (see Fig.~\ref{fig:application} and Fig.~\ref{fig:magneticSystem}).

\textbf{Sliding mode controller with disturbance observer:}
To enable reliable autonomous control under pulsatile flow disturbances and model uncertainties, we developed a sliding mode controller with an integrated disturbance observer (SMC–DOB). While inspired by the general SMC–DOB framework in~\cite{lee2025double}, which was implemented for a different 6-coil electromagnetic system, the controller in this work was newly formulated for the \textit{in vitro} cardiac environment. The sliding surface, switching term, and disturbance observer were designed to account for the millirobot’s hydrodynamic drag, electromagnetic actuation characteristics, and flow disturbances. The controller computes a planar force command $\mathbf{F}_{\mathrm{des}} = [F_{x}, F_{y}]^\top$ to drive the robot from its current position $\mathbf{x}=[x,y]^\top$ to the desired position $\mathbf{x}_d=[x_d,y_d]^\top$, which is mapped to coil currents using Eq.~\eqref{eq:Control_Equation}.

The sliding surface is chosen based on the tracking error $\mathbf{e}=\mathbf{x}_d-\mathbf{x}$ and its derivative $\dot{\mathbf{e}}=\dot{\mathbf{x}}_d-\dot{\mathbf{x}}$, described as
\begin{equation}
\label{eq:sliding_surface}
\mathbf{s} = \gamma \dot{\mathbf{e}} + \lambda \mathbf{e},
\end{equation}
where $\gamma>0$ and $\lambda>0$ determine the convergence rate and damping of the sliding dynamics. In our system, $\gamma$ and $\lambda$ are set to 2 and 1, respectively, for a stable response.

The control law for the desired force includes three terms and is expressed by
\begin{equation}
\label{eq:control_sum}
\mathbf{F}_{\mathrm{des}} = -\hat{\mathbf{d}} + \mathbf{u}_{\mathrm{eq}} + \mathbf{u}_{\mathrm{sw}},
\end{equation}
where $\hat{\mathbf{d}}$ is the disturbance estimate from the disturbance observer (DOB), $\mathbf{u}_{\mathrm{eq}}$ is a model-based equivalent term, and $\mathbf{u}_{\mathrm{sw}}$ is a nonlinear switching term enforcing $\mathbf{s}\to\mathbf{0}$.

The DOB estimates unmodeled drag and flow forces using two low-pass filters:
\begin{align}
\dot{\mathbf{p}} &= -\frac{1}{\eta}\mathbf{W}_p(\mathbf{p}-\dot{\mathbf{e}}),\\
\dot{\mathbf{r}} &= -\frac{1}{\eta}\mathbf{W}_r(\mathbf{r}-\mathbf{F}_{\mathrm{prev}}),
\end{align}
where $\mathbf{p},\mathbf{r}\in\mathbb{R}^2$ are filter states, $\eta>0$ sets the filter bandwidth, $\mathbf{W}_p,\mathbf{W}_r\in\mathbb{R}^{2\times2}$ are diagonal weight matrices, and $\mathbf{F}_{\mathrm{prev}}$ is the previously applied force. The estimated disturbance is
\begin{equation}
\label{eq:dob}
\hat{\mathbf{d}} = m\Big(\frac{1}{\eta}\mathbf{W}_p(\mathbf{p}-\dot{\mathbf{e}})\Big) 
- \mathbf{C}(\dot{\mathbf{x}},\mathbf{V}) - \mathbf{r},
\end{equation}
where $m$ is the robot mass, $\mathbf{V}=[V_x,V_y]^\top$ is the estimated flow velocity, and $\mathbf{C}(\dot{\mathbf{x}},\mathbf{V}) = c_t(\dot{\mathbf{x}}+\mathbf{V})$ represents the modeled viscous and flow-induced drag (see Eq.~\eqref{eq:Drag_force}).

The equivalent term compensates for nominal dynamics and shapes the closed-loop error response:
\begin{equation}
\label{eq:ueq}
\mathbf{u}_{\mathrm{eq}} = m\mathbf{K}_3^{-1}\big(\mathbf{K}_1\dot{\mathbf{e}}+\mathbf{K}_2\mathbf{e}\big) + m\ddot{\mathbf{x}}_d - \mathbf{C}(\dot{\mathbf{x}},\mathbf{V}),
\end{equation}
where $\mathbf{K}_1,\mathbf{K}_2,\mathbf{K}_3\in\mathbb{R}^{2\times2}$ are positive gains (scaled by fluid viscosity) and $\ddot{\mathbf{x}}_d$ is the desired acceleration (set to zero when unknown). The switching term drives the system to the sliding surface while mitigating chattering:
\begin{equation}
\label{eq:usw}
\mathbf{u}_{\mathrm{sw}} = K_4 \tanh\!\left(\frac{\mathbf{s}}{\phi}\right),
\end{equation}
where $K_4>0$ is the switching gain and $\phi>0$ is the boundary-layer thickness. 

\noindent\textbf{Lemma (SMC-DOB stability):}
Let the lumped disturbance $\mathbf{d}(t)$ be bounded and Lipschitz continuous, and let the DOB produce a bounded estimation error $\tilde{\mathbf{d}}=\mathbf{d}-\hat{\mathbf{d}}$ satisfying $\|\tilde{\mathbf{d}}\|\le \delta_{\max}$, where $\delta_{\max}\ge0$ denotes the maximum estimation error bound. If $K_4>\delta_{\max}$, then the error is ultimately uniformly bounded. Furthermore,  if $\delta_{\max}\!\to\!0$, then $\mathbf{s}(t)\!\to\!\mathbf{0}$ and consequently $\mathbf{e}(t)\!\to\!\mathbf{0}$ asymptotically.

\textit{Proof:}
Follows from the  Lyapunov candidate ${V=\tfrac{1}{2}\mathbf{s}^\top\mathbf{s}+\tfrac{1}{2\kappa}\tilde{\mathbf{d}}^\top\tilde{\mathbf{d}}}$ $(\kappa>0)$ and the switching controller in Eq.~\eqref{eq:usw}. If $K_4>\delta_{\max}$, there exists a ball whose radius depends on $\phi$ such that $\dot V\le 0$ for all $\mathbf{s}(t)$ outside this ball. As such, $\|\mathbf{s}(t)\|\to\mathcal{O}(\phi)$. If the DOB converges, then the sliding dynamics converge to the origin, and the error converges to zero asymptotically~\cite{li2012design}.

\textbf{PID baseline controller:} For baseline comparison, a conventional PID controller was implemented to compute planar force commands from the position error $\mathbf{e}=\mathbf{x}_d-\mathbf{x}$. The PID control law is
\begin{equation}
\label{eq:PID_vector}
\mathbf{F}_{\mathrm{PID}}=\mathbf{K}_p\cdot\mathbf{e}+\mathbf{K}_i\cdot\!\int\!\mathbf{e}\,dt+\mathbf{K}_d\cdot\dot{\mathbf{e}},
\end{equation}
where $\mathbf{K}_p,\mathbf{K}_i,\mathbf{K}_d\in\mathbb{R}^{2\times2}$ are diagonal gain matrices. The net force command applied to the robot is
\begin{equation}
\label{eq:PID_net_force}
\mathbf{F}_{\mathrm{des}} = \mathbf{F}_{\mathrm{PID}} - \mathbf{F}_d,\qquad 
\mathbf{F}_d = c_t \mathbf{V},
\end{equation}
where $\mathbf{V}$ is the estimated flow velocity and $c_t$ is the translational damping coefficient from Eq.~\eqref{eq:Drag_coefficient}. Orientation is controlled independently using a unit magnetic field aligned with the desired heading angle $\theta_d$,
\begin{equation}
\label{eq:orientation_field}
\mathbf{B} =
\begin{bmatrix}
\sin\theta_d\\
\cos\theta_d
\end{bmatrix},\qquad \|\mathbf{B}\|=1.
\end{equation}
The desired force $\mathbf{F}_{\mathrm{des}}$ and field $\mathbf{B}$ are converted to coil currents using Eq.~\eqref{eq:Control_Equation}.

\textbf{MPC baseline controller:}
A model predictive control (MPC) framework was implemented as a baseline to compute the desired force $\mathbf{F}_{\mathrm{des}}$ and magnetic field $\mathbf{B}_{\mathrm{des}}$. The robot dynamics are modeled as
\begin{equation}
\label{eq:MPC_function}
M \cdot \mathbf{\dot{v}}(t) + \mathbf{F}_{drag} = \mathbf{F}_{\mathrm{des}}(t),
\end{equation}
where $M$ is the robot mass and $\mathbf{F}_{drag}$ is the drag force from Eq.~\eqref{eq:Drag_force}.

The prediction horizon was set to $N = 20$ with a time step of $\Delta t = 40$ ms. The MPC cost function is
\begin{equation}
J = \sum_{k=0}^{N-1} \left(e_k^T \mathbf{Q} e_k + F_k^T \mathbf{R} F_k\right) + e_N^T \mathbf{P} e_N,
\end{equation}
where $e_k$ and $F_k$ denote the position error and desired force at step $k$, and $\mathbf{Q}$, $\mathbf{R}$, and $\mathbf{P}$ are weighting matrices.

The optimization problem was solved using the Multifrontal Massively Parallel Sparse Direct Solver (MUMPS). The first optimal force input was combined with the magnetic field $\mathbf{B}$ from Eq.~\eqref{eq:orientation_field} and mapped to coil currents using Eq.~\eqref{eq:Control_Equation}.

\section{Experiments and Results}
\noindent
We conducted experiments to evaluate the trajectory tracking performance of the proposed sliding mode controller with disturbance observer (SMC-DOB) in the heart phantom. Control difficulty was systematically varied by changing two disturbance factors: fluid viscosity and flow strength. Tests were performed in high-viscosity (20 cP) and low-viscosity (4.3 cP) fluids, each under static (no flow) and pulsatile flow with peak in-phantom velocities of 7 cm/s and 10 cm/s. These conditions reflect physiologically relevant ranges, as 4.3 cP and 20 cP correspond to near-healthy and elevated blood viscosities~\cite{taco2019association}, and 7–10 cm/s falls within the resting right atrial flow range~\cite{gabe1969measurement}. Proportional–integral–derivative (PID) and model predictive control (MPC) were implemented as baseline controllers. The following subsections first present baseline performance in static fluid, then evaluate controller performance under increasing disturbance levels, and finally assess the contribution of the disturbance observer through ablation experiments.

\subsection{Baseline Trajectory Tracking in Static Fluid}
\noindent
Baseline experiments were first conducted in static fluid to evaluate the trajectory tracking performance of PID, MPC, and SMC-DOB under two viscosity conditions: 20~cP and 4.3~cP. The millirobot was actuated using external magnetic fields (24.05~mT flux density, 13.92~mT/cm gradient) to follow the planned trajectory in Fig.~\ref{fig:astar-path} at a constant speed of 0.5~mm/s, with the field heading fixed at $45^\circ$ relative to the $x$-axis.

For the PID controller, the gains $\boldsymbol{k}_p$, $\boldsymbol{k}_i$, and $\boldsymbol{k}_d$ were set to $5\times10^{-2}$~N$\cdot$m$^{-1}$, $5\times10^{-3}$~N$\cdot$s$^{-1}$$\cdot$m$^{-1}$, and $8\times10^{-3}$~N$\cdot$s$\cdot$m$^{-1}$, respectively. For SMC-DOB, the switching gain $K_4$, disturbance filter bandwidth $\eta$, and boundary-layer thickness $\phi$ were $2.0\times10^{-4}$~N$\cdot$s$^{-2}$, $6.0\times10^{-2}$~s$^{-1}$, and $9.0\times10^{-3}$~m, respectively. Tracking performance was quantified using root mean square error (RMSE), 95th percentile error (P95), and maximum error over three trials per condition (Table~\ref{tab:baseline}).

In 20~cP fluid, SMC-DOB achieved the highest accuracy, with a mean RMSE of 0.49~mm (0.066 body lengths) and maximum error of 1.64~mm. PID also maintained sub-millimeter accuracy, while MPC exhibited larger deviations. In 4.3~cP fluid, both PID and MPC failed to complete the trajectory due to increased disturbance, whereas SMC-DOB successfully tracked it with a mean RMSE of 1.05~mm.

\begin{table}[t]
\centering
\caption{Baseline trajectory tracking in static fluid. Values represent mean $\pm$ standard deviations across three trials per condition.}
\begin{tabular}{lccccc}
\toprule
Controller & Viscosity & RMSE & P95 & Max\\
 & (cP) & (mm) & (mm) & (mm)\\
\midrule
PID        & 20   & $0.82\!\pm\!0.02$ & $1.78\!\pm\!0.05$ & $2.70\!\pm\!0.15$\\
MPC        & 20   & $1.57\!\pm\!0.04$ & $2.77\!\pm\!0.14$ & $4.40\!\pm\!0.38$\\
SMC-DOB    & 20   & $\mathbf{0.49\!\pm\!0.01}$ & $\mathbf{0.86\!\pm\!0.03}$ & $\mathbf{1.64\!\pm\!0.21}$\\
PID        & 4.3  & \multicolumn{3}{c}{\textit{\footnotesize — failed to complete trajectory}} \\
MPC        & 4.3  & \multicolumn{3}{c}{\textit{\footnotesize — failed to complete trajectory}} \\
SMC-DOB    & 4.3  & $\mathbf{1.05\!\pm\!0.02}$ & $\mathbf{1.90\!\pm\!0.05}$ & $\mathbf{3.21\!\pm\!0.23}$\\
\bottomrule
\end{tabular}
\label{tab:baseline}
\vspace{-1.5\baselineskip}
\end{table}

\subsection{Tracking Performance under Moderate Flow Conditions}
\noindent
Trajectory tracking was next evaluated under moderate flow conditions (viscosity: 20~cP; peak in-phantom velocity: 7~cm/s) to compare the proposed SMC–DOB controller with a conventional PID controller. Fluid-drag forces were estimated using time-averaged CFD velocity fields and computed using Eq.~\eqref{eq:Drag_force} and Eq.~\eqref{eq:Drag_coefficient}. The PID controller used the same gains as in static tests, except for an increased derivative gain $\boldsymbol{k}_d = 2.0\times10^{-2}$~N$\cdot$s$\cdot$m$^{-1}$ to provide additional damping against flow-induced oscillations. The SMC–DOB controller was re-tuned with $K_4 = 3.0\times10^{-4}$~N$\cdot$s$^{-2}$, $\eta = 12.0\times10^{-2}$~s$^{-1}$, and $\phi = 18.0\times10^{-3}$~m. Compared to static conditions, these parameters were increased to compensate for higher drag, accelerate disturbance estimation, and suppress chattering while maintaining stability.

\begin{figure}[t]
\centering
\includegraphics[width=0.75\linewidth]{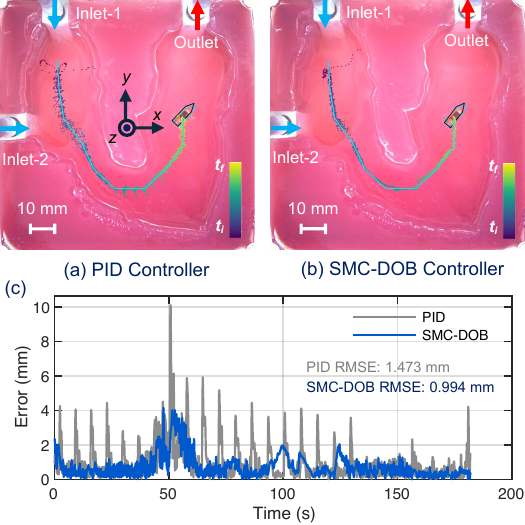}
\caption{Trajectory tracking under moderate flow conditions (20~cP; 7~cm/s peak speed). 
Representative video snapshots show the millirobot’s path, color gradient from start ($t_i$) to end ($t_f$), inside the heart phantom using (a) PID and (b) SMC–DOB controllers. (c) Tracking error over time. 
SMC–DOB maintained a tighter trajectory tracking and reduced error spikes compared to PID, demonstrating improved robustness to flow disturbances.}
\label{fig:PIDvsSMC}
\end{figure}

As shown in Fig.~\ref{fig:PIDvsSMC}, SMC–DOB maintained a tighter trajectory, while PID exhibited larger lateral deviations. Across three trials, PID produced an RMSE of $1.59\!\pm\!0.07$~mm and peak error of $10.47\!\pm\!0.36$~mm, whereas SMC–DOB reduced RMSE to $1.00\!\pm\!0.02$~mm and peak error to $4.36\!\pm\!0.21$~mm. The error profiles in Fig.~\ref{fig:PIDvsSMC}(c) further illustrate this difference: PID generated a sharp error spike of 10.12~mm near $t\!\approx\!50$~s when the robot passed inlet-2 and experienced a sudden flow disturbance. In contrast, SMC–DOB limited the deviation to 4.16~mm by rapidly compensating for the disturbance, demonstrating improved robustness.

\begin{figure}[t]
\centering
\includegraphics[width=0.75\linewidth]{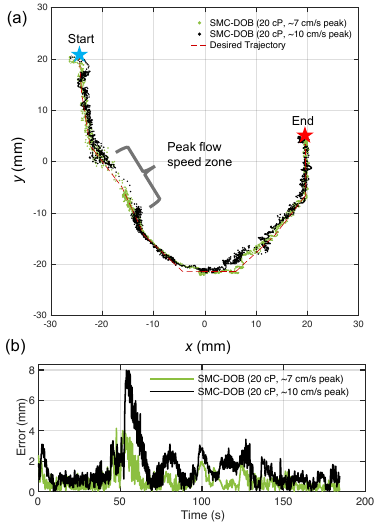}
\caption{Trajectory tracking under moderate flow (20~cP; 7~cm/s peak velocity) and elevated flow (20~cP; 10~cm/s peak velocity) using SMC-DOB controller. (a) Overlaid centroid positions of the millirobot for both conditions in comparison to the desired trajectory in the heart phantom. (b) Tracking error over time for both cases. }
\label{fig:SMC180vs250_error-time}
\vspace{-1\baselineskip}
\end{figure}

\subsection{Robustness to Increased Flow Speed}
\noindent
We next evaluated SMC-DOB robustness under stronger flow disturbances by increasing flow speed while keeping viscosity constant at 20~cP. Pulsatile flow was generated at peak in-phantom velocities of 7~cm/s (moderate flow, 180~mL/min) and 10~cm/s (elevated flow, 250~mL/min). The millirobot followed the same inlet-to-outlet trajectory.

Figure~\ref{fig:SMC180vs250_error-time}(a) shows overlaid trajectories, and Fig.~\ref{fig:SMC180vs250_error-time}(b) shows tracking error over time. A localized high-disturbance region between $y=0$~mm and $y=-10$~mm corresponds to the convergence of inlet streams, producing peak flow velocities. In the heart phantom (Fig.~\ref{fig:magneticSystem}(a)(ii)), this region mimics the narrowed tricuspid valve area, increasing hydrodynamic resistance and disturbance. Under elevated flow, this resulted in gaps in centroid positions in Fig.~\ref{fig:SMC180vs250_error-time}(a), indicating temporary loss of positional hold.

To compensate, the controller was locally retuned when entering this region: the coil current limit was increased by $1.78\times$, and the disturbance filter bandwidth $\eta$ and boundary layer thickness $\phi$ were increased by $1.75\times$ and $1.5\times$, respectively, to improve disturbance rejection while maintaining stability. With these adjustments, the robot completed the trajectory. Across three trials, SMC-DOB maintained RMSE below 2~mm (0.27 body lengths) at 10~cm/s, with a maximum error of $8.2\!\pm\!0.2$~mm, demonstrating reliable autonomous control under substantially stronger pulsatile flow disturbances than previously reported experimental conditions.

\begin{figure}[t]
\centering
\includegraphics[width=0.7\linewidth]{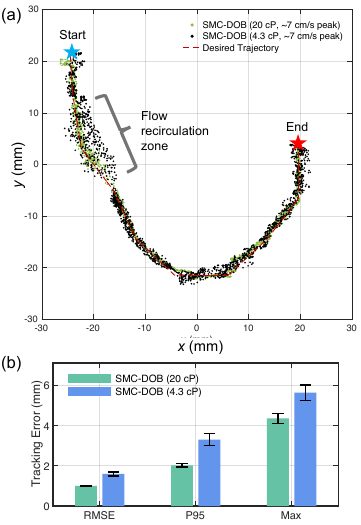}
\caption{Tracking performance of the SMC-DOB controller in 20~cP and 4.3~cP fluids under moderate pulsatile flow (7~cm/s peak speed). 
(a) Overlaid centroid trajectories showing increased oscillations in the low-viscosity case, particularly within the recirculation zone between inlet-1 and inlet-2. 
(b) Comparison of RMSE, P95, and maximum errors across three trials for both viscosities.}
\label{fig:SMC20cPvs4.3cP_trajectoryplot}
\end{figure}

\subsection{Robustness to Viscosity Changes}
\noindent
We further evaluated SMC-DOB robustness to viscosity-induced disturbances by comparing performance in low-viscosity (4.3~cP) and high-viscosity (20~cP) fluids under the same pulsatile flow condition (10~cm/s peak velocity). Due to the pulsatile pump, the flow speed varied over time within the phantom. The millirobot was actuated to autonomously follow the same inlet-to-outlet trajectory.

Figure~\ref{fig:SMC20cPvs4.3cP_trajectoryplot}(a) shows the overlaid trajectories, and Fig.~\ref{fig:SMC20cPvs4.3cP_trajectoryplot}(b) shows the corresponding tracking errors. A localized recirculation zone between $y=10$~mm and $-5$~mm (Fig.~\ref{fig:magneticSystem}(a)(ii)) introduced stronger disturbances at lower viscosity. In 4.3~cP fluid, the millirobot exhibited noticeable oscillations within this region, reflected by the wider spread of centroid positions in Fig.~\ref{fig:SMC20cPvs4.3cP_trajectoryplot}(a). Outside this zone, the trajectory remained closely aligned with the desired path.

Across three trials, the controller achieved an RMSE of $1.60\!\pm\!0.08$~mm, P95 error below 4~mm, and maximum error of $5.64\!\pm\!0.32$~mm in 4.3~cP fluid, compared to an RMSE of $1.00\!\pm\!0.02$~mm and maximum error of $4.36\!\pm\!0.21$~mm in 20~cP fluid. These results demonstrate robust tracking despite viscosity changes and the associated flow disturbances.

\subsection{Effect of Disturbance Observer Ablation}
\begin{figure}[htb]
\centering
\includegraphics[width=0.65\linewidth]{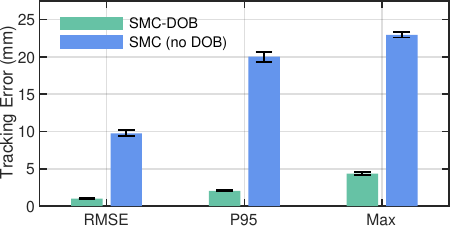}
\caption{Comparison of SMC-DOB and SMC (no DOB) in moderate flow (20~cP, 7~cm/s peak). Bars show RMSE, P95, and maximum tracking errors across three trials.}
\label{fig:dob-ablation}
\vspace{-1\baselineskip}
\end{figure}

\noindent
To evaluate the role of the disturbance observer (DOB), we compared the full SMC-DOB controller with an ablated version (SMC without DOB), in which the observer was disabled while all other parameters were unchanged. Experiments were conducted in the heart phantom under moderate pulsatile flow (20~cP, 7~cm/s peak). As shown in Fig.~\ref{fig:dob-ablation}, removing the DOB resulted in a nearly tenfold increase in RMSE, along with substantially higher P95 and maximum errors. This confirms that the DOB plays a critical role in compensating for flow disturbances and maintaining stable trajectory tracking.

\section{Discussion}\label{sec:dicussion}
\noindent
This work demonstrates autonomous navigation of a magnetic millirobot in a heart phantom under physiologically relevant pulsatile flow. While electromagnetic actuation and visual servoing have been widely studied, prior experimental validation has largely been limited to static or low-disturbance environments. In contrast, pulsatile cardiac flow introduces time-varying disturbances that degrade conventional controllers. The proposed SMC-DOB improves robustness by compensating for modeling uncertainty and transient disturbances, enabling stable tracking under conditions where PID and MPC show degraded performance.

Drag forces were estimated from steady-state CFD simulations under laminar-flow assumptions. Although cardiac flow exhibits transient and complex behavior, the disturbance observer compensates for unmodeled dynamics, as demonstrated experimentally under pulsatile flow conditions. Vision-based localization was used for \textit{in vitro} validation; however, the closed-loop control rate was limited to 8-10 Hz, including image processing and control computation. This rate is within the range of clinical imaging modalities such as X-ray fluoroscopy (7.5-15 Hz)~\cite{abdelaal2014effectiveness} and echocardiography (10-30 Hz)~\cite{mitchell2019guidelines}.

This study has several limitations. Experiments were conducted in a simplified phantom with moderate flow velocities and static path planning. Future work will investigate higher-flow conditions, more complex geometries, integration of clinical imaging, and adaptive planning. Overall, these results demonstrate the feasibility of robust autonomous magnetic millirobot control under pulsatile flow disturbances.

\bibliography{IEEEabrv, references}
\bibliographystyle{IEEEtran}
\end{document}